\pdfoutput=1
\documentclass[11pt]{article}

\usepackage[final]{acl}
\usepackage{booktabs}

\usepackage{times}
\usepackage{latexsym}

\usepackage[T1]{fontenc}
\usepackage[utf8]{inputenc}

\usepackage{microtype}

\usepackage{inconsolata}

\usepackage{graphicx}

\title{When Scale Meets Diversity: Evaluating Language Models on Fine-Grained Multilingual Claim Verification}

\author{
 \textbf{Hanna Shcharbakova\textsuperscript{1}},\\
 \texttt{hash00004@stud.uni-saarland.de} \\
 \textbf{Tatiana Anikina\textsuperscript{2}},
 \textbf{Natalia Skachkova\textsuperscript{2}},
 \textbf{Josef van Genabith\textsuperscript{1,2}}\\
\texttt{\{tatiana.anikina, natalia.skachkova, josef.van\_genabith\}@dfki.de} 
\\
\\
 \textsuperscript{1}Saarland University \\
 \textsuperscript{2}German Research Center for Artificial Intelligence (DFKI)
\\
 }

\begin{document}
\maketitle
\begin{abstract}

The rapid spread of multilingual misinformation requires robust automated fact verification systems capable of handling fine-grained veracity assessments across diverse languages. While large language models have shown remarkable capabilities across many NLP tasks, their effectiveness for multilingual claim verification with nuanced classification schemes remains understudied. We conduct a comprehensive evaluation of five state-of-the-art language models on the X-Fact dataset, which spans 25 languages with seven distinct veracity categories. Our experiments compare small language models (encoder-based XLM-R and mT5) with recent decoder-only LLMs (Llama 3.1, Qwen 2.5, Mistral Nemo) using both prompting and fine-tuning approaches.\footnote{We consider a large language model (LLM) to be any model with more than 1B parameters, and correspondingly, small language model (SLM) to have less than 1B parameters.} Surprisingly, we find that XLM-R (270M parameters) substantially outperforms all tested LLMs (7-12B parameters), achieving 57.7\% macro-F1 compared to the best LLM performance of 16.9\%. This represents a 15.8\% improvement over the previous state-of-the-art (41.9\%), establishing new performance benchmarks for multilingual fact verification. Our analysis reveals problematic patterns in LLM behavior, including systematic difficulties in leveraging evidence and pronounced biases toward frequent categories in imbalanced data settings. These findings suggest that for fine-grained multilingual fact verification, smaller specialized models may be more effective than general-purpose large models, with important implications for practical deployment of fact-checking systems.

\end{abstract}

\section{Introduction}

The rapid spread of misinformation on the internet has become a critical challenge in today's digital age \citep{scheufele2019science, fung2022battlefront}.  With the increasing amount of false information being shared across different languages and platforms, automated fact verification systems have emerged as useful tools for maintaining information reliability. 

The field of automated fact verification has seen significant progress in recent years, particularly with the advent of large language models and transformer-based architectures \citep{guo-etal-2022-survey}. However, most of these advancements have been predominantly focused on English-language content \citep{singhal-etal-2024-multilingual, dmonte2024claim}, creating a significant gap in addressing misinformation in other languages.

Multilingual fact verification presents fundamental challenges for NLP \citep{dmonte2024claim, wang2024monolingualmultilingualmisinformationdetection, zhang-etal-2024-need}, particularly when employing fine-grained classification schemes that better capture the nuanced nature of truth assessment \citep{gupta-srikumar-2021-x, pelrine-etal-2023-towards, Mohtaj}. While existing datasets and approaches employ various classification systems, classification beyond binary (\textit{true}/\textit{false}) and ternary (\textit{true}/\textit{false}/\textit{other}) categories remains 
understudied across multiple languages.

The multi-category nature of this task bears conceptual similarity to Natural Language Inference (NLI) tasks \citep{poliak-etal-2018-hypothesis}, though claim verification differs in its specific objectives. While NLI focuses on determining entailment relationships (\textit{entails, contradicts, neutral}) between premise and hypothesis, our task requires assessing veracity across different distinct truth categories that reflect professional fact-checking standards.

In this work, we examine the performance of diverse model architectures and sizes on multilingual claim verification with fine-grained truth categories. We benchmark language model performance on the X-Fact dataset \citep{gupta-srikumar-2021-x} spanning multiple languages with seven distinct veracity categories, contrasting encoder-based model XLM-R base \citep{conneau-etal-2020-unsupervised}, encoder-decoder architecture mT5 base \citep{xue-etal-2021-mt5}, and recent decoder-only models Llama 3.1 8B \citep{dubey2024llama}, Qwen 2.5 7B \citep{yang2024qwen2}, and Mistral Nemo 12B \citep{mistral}.\footnote{Further details on the specific model versions are provided in Appendix~\ref{app:models}.} For smaller models, we employ standard fine-tuning, while for larger models, we use both parameter-efficient fine-tuning with LoRA \citep{hu2022lowrank} and carefully engineered few-shot prompting approaches. We evaluate models under two conditions: using claims alone and using claims with accompanying evidence text, which allows us to assess both inherent verification capabilities and evidence-augmented reasoning across models using a classification scheme that better reflects the nuanced assessments made by professional fact-checkers. 

Our contributions include:
\begin{itemize}
    \item We conduct comprehensive benchmarking of five state-of-the-art language models on the challenging seven-category multilingual X-Fact dataset, achieving new state-of-the-art results with a 15.8\% improvement in macro-F1 score over previous best performance reported by \citet{gupta-srikumar-2021-x}. We reveal a substantial performance gap between encoder-based and decoder-only architectures despite the latter's greater size and general capabilities.
    \item We provide analysis of model behaviors and error patterns across architectures, identifying several factors that appear to influence multilingual fact verification performance. These observations may help inform future research on verification approaches for diverse languages.
\end{itemize}

\section{Related Work}
\subsection{Multilingual Fact Verification Datasets}

While a substantial portion of fact verification research has centered on English-language content \citep{guo-etal-2022-survey, singhal-etal-2024-multilingual, dmonte2024claim}, several datasets have emerged to address the multilingual dimensions of this challenge. These datasets vary significantly in size, language coverage, and labeling schemes.

% Monolingual datasets are valuable resources for language-specific verification approaches, though they lack cross-lingual frameworks. Following the FEVER \citep{thorne-etal-2018-fever} methodology, researchers have created DanFEVER \citep{norregaard-derczynski-2021-danfever} for Danish with 6K claims and CsFEVER \citep{Ullrich_2023} for Czech with 3K claims. The CHEF dataset \citep{hu-etal-2022-chef} offers 10K claims in Chinese.

Multilingual datasets include FakeCovid \citep{shahi_nandini_2020}, covering 5K claims across 40 languages, and MM-COVID \citep{li2020mmcovidmultilingualmultimodaldata}, which provides 11K articles in English, Spanish, Portuguese, Hindi, French, and Italian. The MultiClaim dataset \citep{pikuliak-etal-2023-multilingual} contains 28K social media posts in 27 languages that can be leveraged for fact verification tasks. FbMultiLingMisinfo \citep{barnabo2022fbmultilingmisinfo} offers 7K news articles spanning 37 languages, while NewsPolyML \citep{Mohtaj} includes 32K claims across English, German, French, Spanish, and Italian. The X-Fact dataset \citep{gupta-srikumar-2021-x} provides 31K claims from fact-checking websites in 25 languages across 11 language families.

% The diversity of labeling schemes across these datasets presents additional challenges for cross-lingual verification. While more nuanced classification systems allow for finer-grained veracity assessments, they also complicate cross-dataset evaluations. 
Labeling approaches range from binary classification \citep{li2020mmcovidmultilingualmultimodaldata, barnabo2022fbmultilingmisinfo} to three-category systems \citep{norregaard-derczynski-2021-danfever, hu-etal-2022-chef, Ullrich_2023} and more complex multi-class schemes including 11 categories in FakeCovid \citep{shahi_nandini_2020}, 4 in NewsPolyML \citep{Mohtaj}, and 7 in X-Fact \citep{gupta-srikumar-2021-x}. The diversity of annotation schemes, while enabling finer-grained veracity assessments, complicates cross-dataset training and evaluation for cross-lingual verification.

\subsection{Methods for Fact Verification}

The task of claim verification has evolved significantly with various methodological approaches emerging to tackle the complexities of determining claim veracity. Transformer-based architectures \citep{devlin-etal-2019-bert} brought substantial advancements to fact verification. \citet{gupta-srikumar-2021-x} evaluated mBERT-based models on the X-Fact dataset spanning 25 languages with 7-way classification. Their best model achieved an F1 score of 41.9\% on the in-domain test set, though performance dropped to 16.2\% F1 on out-of-domain and 16.7\% F1 on zero-shot test sets, highlighting cross-lingual generalization challenges.

Recent research has explored large language models for fact verification using various approaches. For prompting-based methods, \citet{cao2023largelanguagemodelsgood} investigated different prompting strategies for fact-checking, finding that carefully crafted prompts with explicit instructions about expected output formats and task definitions significantly improved performance. \citet{hu2023largelanguagemodelsknow} found that increasing few-shot examples beyond a certain threshold provides substantial gains, suggesting a threshold effect. Self-consistency methods using majority voting from multiple LLM runs improved performance, while self-refinement strategies where models iteratively refine their answers showed gains over standard approaches.

% The number of few-shot examples has varying impacts on LLM's performance on fact verification. Based on English-language data from the Pinocchio-Lite benchmark, \citet{hu2023largelanguagemodelsknow} found modest improvements when increasing from 1-shot to 3-shots (1.5\% gain) but more substantial gains with 12 examples (6.9\% gain), suggesting a threshold effect. Self-consistency methods using majority voting from multiple LLM runs improved performance by an average of 5.0\%, while self-refinement strategies where models iteratively refine their answers showed gains of 6.2\% over standard approaches. 

Chain of Thought (CoT) approaches have shown promising results by enabling LLMs to articulate reasoning processes before reaching conclusions \citep{wei}. \citet{hu2023largelanguagemodelsknow} found that CoT prompting significantly improved performance across all tested models on English data compared to standard prompting.

% \citet{hu2023largelanguagemodelsknow} investigated the effectiveness of different prompting strategies with LLMs for fact verification and found that CoT prompting significantly improved performance across all tested models on English data. They observed a relative increase of 4.3\% in accuracy when using CoT compared to standard prompting. 

\citet{pelrine-etal-2023-towards} compared GPT-4 against traditional approaches across multiple datasets. For binary classification on English LIAR \citep{wang-2017-liar}, GPT-4 variants outperformed traditional models like ConvBERT \citep{NEURIPS2020_96da2f59} and BERT. However, in multi-way classification tasks, performance declined significantly with traditional models like DeBERTa \citep{he2021deberta} showing better results. The same study demonstrated GPT-4's cross-lingual capabilities on CT-FAN-22 \citep{shahi2021overview}, with GPT-4 substantially outperforming RoBERTa-L \citep{liu2019roberta} on English multi-way classification.

% \citet{pelrine-etal-2023-towards} compared GPT-4 against traditional approaches across multiple datasets. For binary classification on English LIAR \citep{wang-2017-liar}, GPT-4 Score Optimized achieved 68.1\% F1, outperforming ConvBERT \citep{NEURIPS2020_96da2f59} (65.8\%) and BERT (64.5\%). However, in 6-way classification, performance declined significantly with GPT-4 Zero-Shot reaching only 25.5\% F1 while DeBERTa \citep{he2021deberta} variant achieved 29.2\% F1. The same study demonstrated GPT-4's cross-lingual capabilities on CT-FAN-22 \citep{shahi2021overview}, with GPT-4 Score Zero-Shot achieving 42.8\% F1 on English and 38.7\% F1 on German data, substantially outperforming RoBERTa-L's \citep{liu2019roberta} 26.8\% on English 4-way classification.

% \citet{cekinel-etal-2024-cross} found that fine-tuning LLaMA-2 models \citep{touvron2023llama2openfoundation} on Turkish language data outperformed cross-lingual transfer methods for fact verification. Their LLaMA-2 13B model fine-tuned with evidence summaries achieved 89.0\% F1 score on binary classification, while cross-lingual prompting with English data showed improvements but proved less effective than Turkish-specific fine-tuning. \citet{Mohtaj} evaluated multiple models on the NewsPolyML dataset spanning five European languages with four veracity categories. Interestingly, mBERT achieved the highest F1 score of 75.1\%, suggesting that model size does not necessarily correlate with performance in multilingual fact verification tasks.

\citet{cekinel-etal-2024-cross} found that fine-tuning LLaMA-2 models \citep{touvron2023llama2openfoundation} on Turkish language data outperformed cross-lingual transfer methods for fact verification. Their fine-tuned model achieved strong performance on binary classification, while cross-lingual prompting with English data showed improvements but proved less effective than language-specific fine-tuning. \citet{Mohtaj} evaluated multiple models on the NewsPolyML dataset spanning five European languages with four veracity categories. Interestingly, mBERT achieved the highest performance, suggesting that model size does not necessarily correlate with performance in multilingual fact verification tasks.

\section{Dataset}

For our experiments, we use the X-Fact dataset \citep{gupta-srikumar-2021-x}. This dataset was selected due to several advantages over other multilingual fact verification resources. X-Fact encompasses a broad range of topics from verified fact-checking websites, making it more representative of real-world misinformation challenges compared to specialized datasets like FakeCovid \citep{shahi_nandini_2020} that focus solely on COVID-19 related claims. Unlike datasets derived from social media platforms such as FbMultiLingMisinfo \citep{barnabo2022fbmultilingmisinfo}, X-Fact provides ready-to-use data without requiring access to platform-specific APIs, ensuring reproducibility of research findings.

X-Fact comprises 31,189 claims across 25 languages from 11 language families, including Indo-European, Afro-Asiatic, Austronesian, Kartvelian, Dravidian, and Turkic. The dataset was carefully constructed by identifying reliable fact-checking sources from the International Fact-Checking Network\footnote{https://www.poynter.org/ifcn/.} and Duke Reporter's Lab\footnote{https://reporterslab.org/.}, excluding websites that conduct fact-checks in English to avoid overlap with existing datasets. Each claim in X-Fact is accompanied by up to 5 pieces of evidence extracted from fact-checking articles, with an average of 4.75 non-empty evidence pieces per claim. The dataset also includes valuable metadata such as the language of the claim and evidence, the fact-checking site where the claim was derived from, links to the evidence where they were published, claim date, review date, and claimant information. Examples of claims, corresponding evidence, and associated metadata can be found in Appendix~\ref{app:claim}.

To ensure consistent evaluation across different fact-checking standards, the dataset employs a standardized seven-label classification scheme:\textit{ true, mostly true, partly true/misleading, mostly false, false, complicated/hard to categorize}, and \textit{other}. This fine-grained approach provides a more nuanced assessment of claim veracity compared to less fine-grained classification schemes used in many other datasets.

The dataset is divided into multiple subsets designed to evaluate different aspects of model performance (see Table \ref{tab:subsets}). The training set contains 19,079 claims across 13 languages, while the development set comprises 2,535 claims spanning 13 languages. For testing, X-Fact provides three separate subsets: an in-domain test set with 3,826 claims from the same languages and sources as the training data; an out-of-domain test set containing 2,368 claims from the same languages but different sources; and a zero-shot test set featuring 3,381 claims from 12 languages not present in the training data. This evaluation framework supports a thorough assessment of models' generalization capabilities across both domains and languages.

\begin{table}[h]
  \centering
  \begin{tabular}{l|c|c}
    \hline
    \textbf{Dataset Subset}  & \textbf{\# Claims} & \textbf{\# Languages} \\
    \hline
    Training & 19079 & 13 \\
    Development & 2535 & 13 \\
    In-domain & 3826 & 13 \\
    Out-of-domain & 2368 & 5 \\
    Zero-shot & 3381 & 12 \\\hline
  \end{tabular}
  \caption{Overview of the X-Fact dataset subsets.}
  \label{tab:subsets}
\end{table}

\begin{figure}[h]
  \includegraphics[width=\columnwidth]{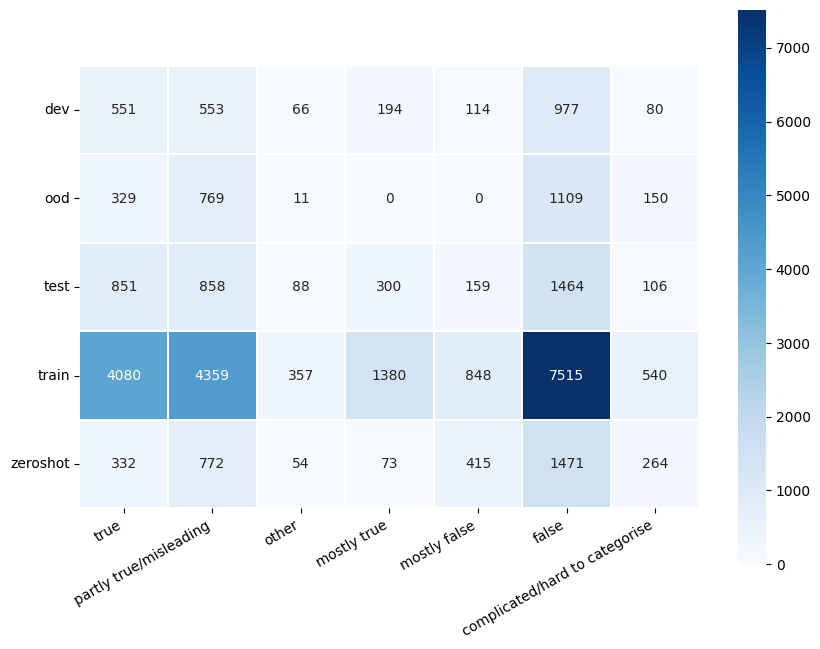}
  \caption{Distribution of data in X-Fact by label across subsets.}
  \label{fig:label}
\end{figure}

The label distribution in the X-Fact exhibits significant variation across all subsets (see Figure \ref{fig:label}). The \textit{false} label dominates the training set with 7,515 instances (39.4\%), followed by \textit{partly true/misleading} with 4,359 instances (22.8\%). The least represented label is \textit{other} with only 576 instances (1.9\%). 

\begin{figure*}[h]
  \includegraphics[width=\textwidth]{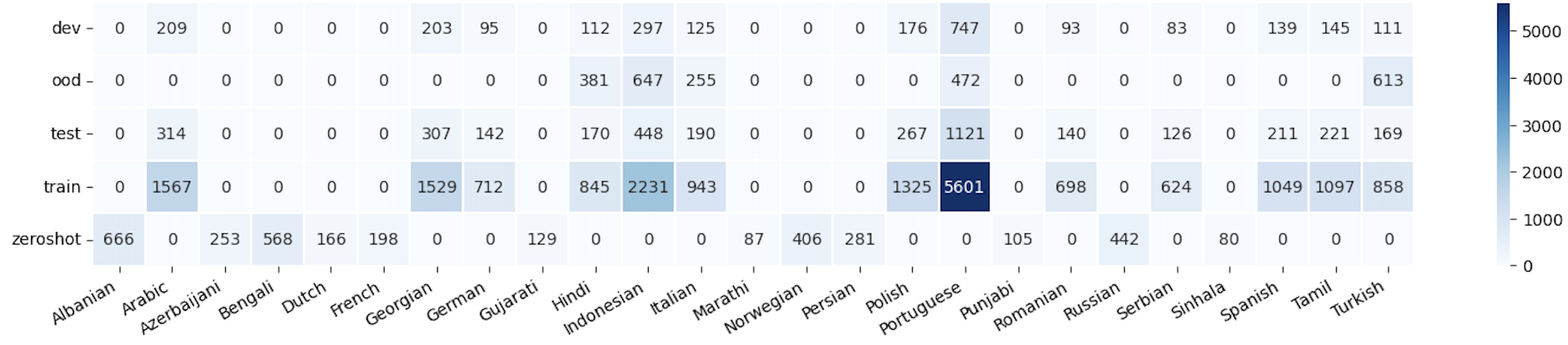}
  \caption{Distribution of data in X-Fact by language across subsets.}
  \label{fig:lang}
\end{figure*}

The language distribution also shows substantial variation across different subsets (see Figure \ref{fig:lang}). Portuguese dominates the training set with 5,601 claims (29.4\%), followed by Indonesian with 2,231 claims (11.7\%) and Arabic with 1,567 claims (8.2\%). Serbian has the lowest representation with only 624 claims (3.3\%). 

These imbalances may potentially impact model learning, particularly for cross-lingual transfer, and present additional challenges for models to learn fine-grained veracity categories.

\section{Experiments}

\subsection{Experimental Setup}

Our evaluation focuses on benchmarking different language model architectures on the multilingual fact verification task, using X-Fact's seven-category classification scheme across multiple languages. We evaluate both small language models (SLMs, <1B parameters) and large language models (LLMs, >1B parameters) to determine their relative effectiveness for fine-grained multilingual verification. Table \ref{tab:models} provides an overview of the models evaluated in this study.

We selected these models based on their strong multilingual capabilities and architectural diversity. XLM-R was chosen for its robust pre-training on 100 languages and encoder-only architecture that has proven effective for classification tasks. MT5 represents the encoder-decoder paradigm, offering a different architectural approach while maintaining strong multilingual capabilities across 101 languages. For LLMs, we selected Llama 3.1 8B, Qwen 2.5 7B, and Mistral Nemo 12B to represent state-of-the-art decoder-only architectures with varying degrees of multilingual support.

We prioritized open-source models with moderate parameter sizes to ensure reproducibility and facilitate deployment in resource-constrained environments. This selection allows us to evaluate whether sophisticated reasoning in current LLMs transfers effectively to multilingual fact verification compared to smaller, specialized architectures like XLM-R.

\begin{table}[h]
\centering
\begin{tabular}{p{1.6cm}|p{1cm}|p{1.2cm}|p{2.3cm}}
\hline
\textbf{Model} & \textbf{\# Par.} & \textbf{\# Lang.} & \textbf{Architecture} \\
\hline
XLM-R & 270 M & 100 & Encoder-only \\\hline
mT5 & 580 M & 101 & Encoder-decoder \\\hline
Llama 3.1 & 8 B & 8 & Decoder-only \\\hline
Qwen 2.5 & 7 B & 29 & Decoder-only \\\hline
Mistral Nemo & 12 B & 11 & Decoder-only \\\hline
\end{tabular}
\caption{Models evaluated on multilingual fact verification using the X-Fact dataset. \# Par. is the number of parameters and \# Lang. is the number of languages supported by each model.}
\label{tab:models}
\end{table}

For the SLMs, we performed fine-tuning experiments, while for LLMs, we explored both direct prompting and parameter-efficient fine-tuning using LoRA. The models' implementation details can be found in the Appendix \ref{app:hypers}.

\subsection{Small Language Models Experiments}

For XLM-R and mT5, we conducted two types of fine-tuning experiments:

\begin{itemize}
    \item \textbf{Full Model Fine-tuning:} We performed complete fine-tuning of the models, allowing all parameters to be updated during training.
    \item \textbf{Classification Head Fine-tuning:} We fine-tuned only the classification head while keeping the base model frozen.
\end{itemize}

For both approaches, we provided the models with the claim text and evidence as input. We did not conduct experiments using only claim text without evidence, as preliminary experiments confirmed the X-Fact paper's finding that claim-only setups yield worse performance.

\subsection{Large Language Models Experiments}

% For LLMs, we explored both few-shot prompting and parameter-efficient fine-tuning approaches. We evaluated each LLM in two input configurations: (1) claim-only, providing only the claim text; and (2) claim with evidence, providing both claim and evidence text. 

% For prompting experiments, we developed carefully engineered 7-shot prompts where the number of examples corresponds to the seven veracity categories in our classification scheme. Each prompt included clear instructions, defined the expected output format, and provided descriptions of the seven veracity categories. The optimized prompt is provided in the Appendix \ref{app:prompt}.

% We implemented parameter-efficient fine-tuning using LoRA, targeting both attention components and feed-forward layers. We tested each LLM in four distinct modes: inference with claim-only input, inference with claim and evidence, LoRA fine-tuning with claim-only input, and LoRA fine-tuning with claim and evidence. Further details can be found in Appendix \ref{app:hypers}.

For LLMs, we explored both few-shot prompting and parameter-efficient fine-tuning approaches. We evaluated each model in two input configurations: (1) claim-only, providing only the claim text; and (2) claim with evidence, providing both claim and evidence text. Our experimental setup included the following approaches:

\begin{itemize}
    \item \textbf{Few-shot prompting:} We developed 7-shot prompts containing examples for each veracity category to guide prediction without training. Each prompt included clear instructions, category definitions, and was tested in both claim-only and claim+evidence variants.
    \item \textbf{LoRA fine-tuning:} We implemented parameter-efficient fine-tuning using LoRA for both claim-only and claim+evidence configurations. 
\end{itemize}

The optimized prompt template is provided in the Appendix \ref{app:prompt}. 

\section{Results}

% This section presents our experimental findings on multilingual fact verification, comparing the performance of different model architectures on the X-Fact dataset.

\subsection{SLMs Performance}

Table \ref{tab:slm_results} presents the macro-F1 scores for small language models across three evaluation subsets. XLM-R with full fine-tuning achieves the highest performance with 57.7\% macro-F1 on the test set, substantially outperforming the previous state-of-the-art mBERT baseline (41.9\%) by 15.8\% reported in \citet{gupta-srikumar-2021-x}. XLM-R also demonstrates superior cross-domain and cross-lingual generalization, maintaining relatively strong performance across all evaluation subsets. 

\begin{table}[h]
  \centering
  \begin{tabular}{l|c|c|c}
    \hline
    \textbf{Model} & \textbf{Test} & \textbf{OOD} & \textbf{Zero-shot} \\
    \hline
    mBERT (baseline) & 41.9 & 16.2 & 16.7 \\
    XLM-R frozen & 51.4 & 40.8 & 41.3 \\
    XLM-R & \textbf{57.7} & \textbf{47.6} & \textbf{43.2} \\
    mT5 & 47.6 & 22.2 & 19.2 \\
    \hline
  \end{tabular}
  \caption{SLMs performance on the X-Fact dataset (macro-F1 scores). XLM-R frozen refers to fine-tuning the classification head only. mBERT performance is derived from \citep{gupta-srikumar-2021-x}.}
  \label{tab:slm_results}
\end{table}

MT5 reaches 47.6\% macro-F1 on the test set but shows poor generalization to out-of-domain (22.2\%) and zero-shot (19.2\%) scenarios. The performance gap between XLM-R and mT5 widens significantly on these evaluation sets, indicating that XLM-R's encoder-only architecture may be better suited for multilingual fact verification tasks.

\subsection{LLMs Performance}

Table \ref{tab:llm_results} presents LLMs' results across different configurations. Despite their significantly larger size (7-12B parameters), all LLMs substantially underperform compared to SLMs. The best LLM configuration (Qwen claim-only fine-tuning) achieves only 16.9\% macro-F1 on the test set - 40.8\% points below XLM-R. For visualizations of models' performance across different evaluation subsets, refer to Appendix \ref{app:all}.

\begin{table*}[t]
  \centering
  \begin{tabular}{p{1.5cm}|p{1cm}p{1cm}|p{1cm}p{1cm}|p{1cm}p{1cm}|p{1cm}p{1cm}}
    \hline
{\textbf{Method}} & \multicolumn{4}{c|}{\textbf{Few-shot}} & \multicolumn{4}{c}{\textbf{LoRA-based Finetune}} \\
    \cline{2-9}
    & \multicolumn{2}{c|}{\textbf{Claim+Evidence}} & \multicolumn{2}{c|}{\textbf{Claim Only}} & \multicolumn{2}{c|}{\textbf{Claim+Evidence}} & \multicolumn{2}{c}{\textbf{Claim Only}} \\
    \cline{2-9}
    & \textbf{macro} & \textbf{micro} & \textbf{macro} & \textbf{micro} & \textbf{macro} & \textbf{micro} & \textbf{macro} & \textbf{micro} \\
    \hline
    \multicolumn{9}{l}{\textit{Qwen 2.5}} \\
    \hline
    Test & 12.7 & 24.9 & 11.4 & 18.6 & 15.9 & \textbf{39.5} & \textbf{16.9} & 29.6 \\
    OOD & 13.0 & 29.6 & 11.2 & 27.4 & \textbf{15.1} & \textbf{47.1} & 11.1 & 31.3 \\
    Zero-shot & 10.9 & 18.9 & 12.9 & 23.9 & \textbf{15.4} & \textbf{35.8} & 11.7 & 24.5 \\
    \hline
    \multicolumn{9}{l}{\textit{Mistral Nemo}} \\
    \hline
    Test & \textbf{14.8} & 30.8 & 8.5 & 23.4 & 14.6 & \textbf{31.9} & 10.3 & 20.2 \\
    OOD & \textbf{16.1} & \textbf{42.6} & 9.7 & 36.6 & 12.1 & 34.2 & 9.6 & 27.1 \\
    Zero-shot & \textbf{15.1} & 28.7 & 10.6 & \textbf{29.6} & 12.9 & 25.7 & 8.2 & 15.6 \\
    \hline
    \multicolumn{9}{l}{\textit{Llama 3.1}} \\
    \hline
    Test & 14.0 & \textbf{32.0} & 10.8 & 18.4 & 14.3 & 27.6 & \textbf{15.5} & 30.5 \\
    OOD & 13.3 & \textbf{41.2} & 8.7 & 21.2 & 11.2 & 27.1 & \textbf{13.5} & 33.2 \\
    Zero-shot & \textbf{12.9} & \textbf{30.1} & 8.7 & 17.6 & 9.6 & 17.5 & 12.1 & 29.4 \\
    \hline
  \end{tabular}
  \caption{LLMs performance on the X-Fact dataset (macro-F1 and micro-F1 scores). Bold values indicate the highest macro- and micro-F1 scores for each model-subset combination.}
  \label{tab:llm_results}
\end{table*}

Among the LLMs, Qwen 2.5 consistently demonstrates the best performance across most configurations. The model achieves its highest macro-F1 score of 16.9\% with claim-only fine-tuning on the test set, compared to 15.9\% with claim+evidence fine-tuning and 11.4\%-12.7\% with few-shot prompting. LoRA-based fine-tuning consistently improves performance over few-shot inference across all models and configurations, with Qwen 2.5 showing the largest gains. 
% (from 11.4\% to 16.9\% in claim-only configuration on the test set).

The impact of adding evidence to claims varies significantly across models and methods. For Qwen 2.5, fine-tuning with claim-only (16.9\%) outperforms claim+evidence (15.9\%) on the test set, showing a consistent pattern across all evaluation sets. In contrast, Mistral Nemo generally performs better with claim+evidence input in few-shot settings (14.8\% vs 8.5\% on test set) but shows mixed results with fine-tuning. Llama 3.1 demonstrates the most inconsistent performance across different configurations. While it achieves reasonable performance on the test set (15.5\% macro-F1 with claim-only fine-tuning), it shows the largest performance drop on the zero-shot set, with the worst configuration (claim+evidence fine-tuning) falling to 9.6\% macro-F1.

% Mistral Nemo shows competitive performance, particularly on the out-of-domain set where it achieves its highest score of 16.1\% macro-F1 with claim+evidence few-shot prompting, compared to 12.1\% with fine-tuning in the same configuration. However, like other LLMs, it shows performance degradation on the zero-shot set, reaching only 15.1\% macro-F1 in its best configuration (claim+evidence few-shot).

% Llama 3.1 demonstrates the most inconsistent performance across different configurations. While it achieves reasonable performance on the test set (15.5\% macro-F1 with claim-only fine-tuning), it shows the largest performance drop on the zero-shot set, with the worst configuration (claim+evidence fine-tuning) falling to 9.6\% macro-F1, making it the only model to drop below 10\% in any configuration.

\textbf{LoRA Fine-tuning and Few-shot Prompting.} Fine-tuning consistently improves performance over few-shot prompting across all models. Qwen shows the most substantial improvement (from 12.7\% in few-shot with claim+evidence setting to to 15.9\% in fine-tuning with claim+evidence setting) on the test set. Mistral Nemo shows minimal differences between methods, with some configurations favoring few-shot prompting (16.1\% vs 12.1\% on out-of-domain with claim+evidence). Llama 3.1 generally benefits from fine-tuning, improving from 10.8\% to 15.5\% in the claim-only configuration on the test set.

\textbf{Performance Across Evaluation Subsets.} All models show declining performance from test to out-of-domain and zero-shot sets when fine-tuning. Qwen 2.5 maintains stable performance, with the sharpest drop by 5.2\% from test to zero-shot. Mistral Nemo shows the least variation, performing best on out-of-domain (16.1\%). Llama 3.1 exhibits the largest degradation, dropping from 15.5\% on test to 9.6\% on zero-shot in comparable configurations. Refer to the Appendix \ref{app:set} for visualizations comparing LLMs performance across these evaluation subsets. For a combined view of claim+evidence configurations across all LLMs, refer to Appendix \ref{app:consol}, which directly compares the macro-F1 scores across evaluation subsets and highlights the best performing method for each model.

\textbf{Macro vs. Micro F1 Score.} The substantial gap between micro- and macro-F1 scores is consistent across all LLMs, with the largest gaps observed in fine-tuning configurations. Qwen 2.5 achieves 39.5\% micro-F1 compared to 15.9\% macro-F1 in its claim+evidence fine-tuning on the test set, a gap of 23.6\%. Similarly, Mistral Nemo shows a 17.2\% gap in its claim+evidence fine-tuning configuration on the test set. 

Few-shot configurations generally show smaller gaps. For instance, Qwen's few-shot claim+evidence on the test set shows an 12.2\% gap, while its fine-tuning equivalent shows a 23.6\% gap. This pattern holds across all models where fine-tuning configurations consistently exhibit gaps ranging from 15 to 32 percentage points, while few-shot configurations typically show gaps between 8 to 20 percentage points. The visualizations depicting the performance gaps between macro- and micro-F1 scores across LLMs can be found in Appendix \ref{app:f1}.

\section{Discussion}

Our comprehensive evaluation across 25 languages reveals several important findings that advance our understanding of how different architectures handle fine-grained veracity classification across languages.

\textbf{Performance Gap Between Model Types.} The most striking finding is XLM-R's superiority over all tested LLMs, achieving 57.7\% macro-F1 compared to the best LLM performance of 16.9\% from Qwen 2.5. This performance difference is particularly noteworthy given that LLMs contain many more parameters than XLM-R. XLM-R was pre-trained on 100 languages using a masked language modeling objective that may align well with classification tasks, whereas LLMs use next-token prediction objectives optimized for text generation. These differences in pre-training approaches and objectives may contribute to the observed performance gap.

While our comparison involves different training methodologies (full fine-tuning for SLMs versus LoRA for LLMs), it is important to note that even when comparing more similar approaches, substantial performance gaps persist. Our frozen XLM-R configuration, which only updates the classification head similar to LoRA's parameter-efficient approach, still achieves 51.4\% macro-F1 compared to the best LLM performance of 16.9\%. This suggests that the performance differences extend beyond training methodology. Future work should include detailed per-label performance analysis to better understand model biases and identify which veracity categories prove most challenging across different architectures.

% \textbf{Evidence Integration Patterns.} A clear pattern emerges in how LLMs handle evidence: surprisingly, incorporating additional evidence often does not enhance performance and can even lead to worse results. For instance, Qwen's claim-only fine-tuning (16.9\%) outperforms its claim+evidence configuration (15.9\%). This pattern persists across all LLMs, suggesting systematic difficulties in leveraging additional context for verification decisions.

% This finding is particularly significant because evidence-based reasoning is fundamental to reliable fact verification. The fact that simply providing claims yields better results than including supporting evidence indicates that current LLMs may not be effectively utilizing the additional information or may be getting confused by the increased input complexity.

\textbf{Evidence Integration Patterns.} A clear pattern emerges in how LLMs handle evidence: surprisingly, incorporating additional evidence often does not enhance performance and can even lead to worse results. For instance, Qwen's claim-only fine-tuning (16.9\%) outperforms its claim+evidence configuration (15.9\%). This pattern persists across all Llama 3.1 configurations, suggesting systematic difficulties in leveraging additional context for verification decisions.

We hypothesize several factors that may contribute to this counterintuitive finding. First, the architectural limitations of decoder-only LLMs may hinder effective evidence integration. Unlike XLM-R's bidirectional attention that allows simultaneous consideration of all evidence elements against all claim components, LLMs' autoregressive attention can only consider previous tokens. This sequential processing creates a tendency to forget or ignore earlier information as sequences become longer, making balanced evidence evaluation more challenging.

Second, our input formatting may have contributed to this issue. While we used clear demarcation between claims and evidence in our prompts (as shown in Appendix \ref{app:prompt}), we did not implement more sophisticated structuring techniques that might have helped LLMs better distinguish and compare these elements. Context window size was treated as a hyperparameter in our experiments, with LLMs tested at both 2048 and 4096 tokens, while SLMs were evaluated with context windows ranging from 256 to 512 tokens. With evidence pieces having median lengths of 25-35 words each and approximately 4.75 pieces per claim on average, the evidence was fully accommodated within the context windows of all models. Therefore, evidence truncation was not a contributing factor to the observed performance patterns.

This finding is particularly significant because evidence-based reasoning is fundamental to reliable fact verification. The fact that simply providing claims yields better results than including supporting evidence indicates that current LLMs may not be effectively utilizing the additional information or may be getting confused by the increased input complexity.

\textbf{Fine-Grained Classification Challenges.} The severe data imbalances in X-Fact likely contributes to the observed performance patterns. The dominance of \textit{false} and \textit{partly true/misleading} categories creates a challenging environment for models to learn effective representations for less frequent but equally important categories. This imbalance effect is aggravated in the seven-category setting, where models must not only distinguish between \textit{true} and \textit{false} but also navigate subtle gradations of partial truth. Furthermore, the language distribution imbalance (Portuguese comprising 29.4\% of training data while Serbian represents only 3.3\%) likely impacts cross-lingual performance. Models may develop language-specific biases that hinder their ability to generalize across languages, particularly to those underrepresented in the training data.

% To assess the impact of label granularity, we conducted additional experiments with a coarse-grained classification scheme combining the seven original categories into three broader classes: \textit{true} (combining \textit{true}, \textit{mostly true}, and \textit{partly true/misleading}), \textit{false} (combining \textit{false} and \textit{mostly false}), and \textit{other} (combining \textit{complicated/hard to categorize}, and \textit{other}). These experiments revealed substantial performance improvements across LLMs. For several language-model combinations, the performance more than doubled, with Llama 3.1 showing a remarkable 105\% improvement on certain language subsets. These experiments revealed substantial performance improvements across all models.

The substantial disparity between micro- and macro-F1 scores across all LLMs reveals critical limitations in handling nuanced veracity categories. The micro-F1 scores being consistently higher than macro-F1 scores confirms that performance is driven primarily by accuracy on frequent categories, while rare categories remain poorly predicted. This pattern is particularly pronounced in LLMs, suggesting they may be more influenced by biases in the training data than the fine-tuned XLM-R.

\textbf{Cross-Lingual and Cross-Domain Generalization.} Performance degradation across evaluation subsets is consistent across all models but varies in magnitude. XLM-R demonstrates the most robust cross-lingual transfer, while LLMs show steeper drops. The relatively stable performance of XLM-R on unseen languages suggests that its multilingual pre-training provides effective cross-lingual representations for fact verification task. The sharper declines observed in LLMs may indicate that their multilingual capabilities are less robust when faced with languages not well represented in their training data or when transferring across different fact-checking domains.

Even when comparing XLM-R's frozen configuration (which only updates the classification head, similar to LoRA's parameter-efficient approach), we still observe substantial outperformance over LLMs (51.4\% vs 16.9\% best LLM performance). This suggests that the performance differences may stem not only from the fine-tuning methodology but also from other factors such as architectural advantages of encoder-based models for this specific task or the amount and quality of the pre-training data available in different languages.

\section{Conclusion and Future Work}

This work presents a comprehensive evaluation of diverse language model architectures (small and large; encoder, encoder-decoder, and decoder-only) on multilingual fact verification using the challenging seven-category X-Fact dataset. Our findings reveal several key insights that advance understanding of how different models handle fine-grained veracity classification across languages.

Fully fine-tuned XLM-R emerges as the clear winner, achieving 57.7\% macro-F1 on the test set – a 15.8\% improvement over previous state-of-the-art. Despite having significantly fewer parameters, XLM-R substantially outperforms all tested LLMs, with the best LLM (Qwen 2.5) reaching only 16.9\% macro-F1. The magnitude of this performance gap persists even when comparing lightweight fine-tuning approaches (e.g., frozen XLM-R with a trained classification head: 51.4\% vs best LLM: 16.9\%), suggesting that factors beyond training methodology contribute to the observed differences. However, the exact nature of these factors requires further investigation.

Our analysis reveals problematic patterns in LLM behavior, particularly their inability to effectively utilize additional evidence. Models often perform worse when provided with claim-evidence pairs compared to claims alone, indicating systematic challenges in leveraging external information for verification decisions. This limitation is particularly problematic given that evidence-based reasoning is fundamental to reliable fact-checking, though the underlying causes of this behavior need deeper exploration.

The significant disparity between micro- and macro-F1 scores across the models reveals the challenge of handling imbalanced datasets with fine-grained categories. Models tend to learn shortcuts based on frequent categories while struggling with rare but equally important veracity labels. This bias appears more pronounced in LLMs, indicating they may be more vulnerable to dataset imbalances than smaller models that have been carefully fine-tuned.

These findings have important implications for the development of multilingual fact verification systems. While LLMs show promise for many NLP tasks, our results suggest that for fine-grained fact verification across languages, smaller specialized models may provide better performance while requiring fewer computational resources.

\section{Limitations}

Our study has several limitations that should be considered when interpreting the results.

\textbf{Training Methodology Differences.} Our comparison involves fundamentally different training approaches: XLM-R undergoes full fine-tuning with all parameters being updated, while LLMs utilize LoRA that freezes the majority of the original model parameters. This methodological difference could significantly impact the ability of LLMs to adapt to the specific task requirements and may partially explain the observed performance gaps.

\textbf{Prompt Engineering Constraints.} Our prompt engineering approach may not be equally optimal across all languages in our multilingual evaluation. While we developed carefully engineered 7-shot prompts with examples balanced across the seven veracity categories, our prompt design focused primarily on ensuring representative coverage of each label rather than optimizing for linguistic diversity. This approach may have favored certain languages or language families that were better represented in our example selection. Language-specific prompt optimization could potentially narrow the performance gap, though this would require substantial additional engineering effort for each target language.

\textbf{Evidence Interpretation Limitations.} Given the relatively small performance differences between claim-only and claim+evidence configurations, we cannot definitively conclude that LLMs are incapable of evidence utilization. The limited performance gap may simply reflect the inherent difficulty of the task or limitations in our evaluation approach. It's possible that with more sophisticated prompting strategies, larger datasets, or alternative evidence presentation formats, LLMs might demonstrate improved evidence integration capabilities. The evidence quality in the X-Fact dataset may also play a role, as analysis reveals that search snippets may not always contain sufficient information for accurate verification \citep{gupta-srikumar-2021-x}.

\textbf{Practical Computing Considerations.} Our comparison between fully fine-tuned XLM-R and LoRA-adapted LLMs reflects realistic scenario with limited computational resources. Full fine-tuning of billion-parameter models requires substantial computational resources that are often prohibitive for many researchers. In contrast, parameter-efficient methods like LoRA can be applied with modest computational resources, making them the more practical choice for deploying large models. This comparison addresses a critical question: given realistic computational constraints, which approach provides better performance for multilingual fact verification? Our results demonstrate that a smaller, fully fine-tuned model can significantly outperform much larger models adapted with parameter-efficient methods, suggesting that for specific tasks like multilingual fine-grained verification, specialized smaller models may be preferable to general-purpose large models.

\textbf{Output Analysis and Reproducibility.} To enhance reproducibility and enable further investigation of the observed performance patterns, we make our LLM outputs available in the repository\footnote{https://github.com/Aniezka/xfact-fever.}, including detailed predictions and model responses. A comprehensive analysis of these outputs, including confusion matrices and detailed error patterns that could reveal potential parsing issues or systematic biases, represents important future work that could provide deeper insights into the substantial performance differences observed between model architectures.

\section{Acknowledgments}

This project was supported by the German Federal Ministry of Research, Technology and Space (BMFTR) as part of the project TRAILS (01IW24005), which provided funding for Josef van Genabith and Tatiana Anikina. This work was also co-funded by the Erasmus Mundus Masters Programme in Language and Communication Technologies (EU grant no. 2019-1508).

\bibliography{custom}

\begin{thebibliography}{36}
\providecommand{\natexlab}[1]{#1}

\bibitem[{Barnab{\`o} et~al.(2022)Barnab{\`o}, Siciliano, Castillo, Leonardi, Nakov, Da~San~Martino, and Silvestri}]{barnabo2022fbmultilingmisinfo}
Giorgio Barnab{\`o}, Federico Siciliano, Carlos Castillo, Stefano Leonardi, Preslav Nakov, Giovanni Da~San~Martino, and Fabrizio Silvestri. 2022.
\newblock Fbmultilingmisinfo: Challenging large-scale multilingual benchmark for misinformation detection.
\newblock In \emph{2022 International Joint Conference on Neural Networks (IJCNN)}, pages 1--8. IEEE.

\bibitem[{Cao et~al.(2023)Cao, Wei, Chen, Zhou, and Hu}]{cao2023largelanguagemodelsgood}
Han Cao, Lingwei Wei, Mengyang Chen, Wei Zhou, and Songlin Hu. 2023.
\newblock \href {https://arxiv.org/abs/2311.17355} {Are large language models good fact checkers: A preliminary study}.
\newblock \emph{Preprint}, arXiv:2311.17355.

\bibitem[{Cekinel et~al.(2024)Cekinel, Karagoz, and {\c{C}}{\"o}ltekin}]{cekinel-etal-2024-cross}
Recep~Firat Cekinel, Pinar Karagoz, and {\c{C}}a{\u{g}}r{\i} {\c{C}}{\"o}ltekin. 2024.
\newblock \href {https://aclanthology.org/2024.lrec-main.368/} {Cross-lingual learning vs. low-resource fine-tuning: A case study with fact-checking in {T}urkish}.
\newblock In \emph{Proceedings of the 2024 Joint International Conference on Computational Linguistics, Language Resources and Evaluation (LREC-COLING 2024)}, pages 4127--4142, Torino, Italia. ELRA and ICCL.

\bibitem[{Conneau et~al.(2020)Conneau, Khandelwal, Goyal, Chaudhary, Wenzek, Guzm{\'a}n, Grave, Ott, Zettlemoyer, and Stoyanov}]{conneau-etal-2020-unsupervised}
Alexis Conneau, Kartikay Khandelwal, Naman Goyal, Vishrav Chaudhary, Guillaume Wenzek, Francisco Guzm{\'a}n, Edouard Grave, Myle Ott, Luke Zettlemoyer, and Veselin Stoyanov. 2020.
\newblock \href {https://doi.org/10.18653/v1/2020.acl-main.747} {Unsupervised cross-lingual representation learning at scale}.
\newblock In \emph{Proceedings of the 58th Annual Meeting of the Association for Computational Linguistics}, pages 8440--8451, Online. Association for Computational Linguistics.

\bibitem[{Daniel~Han and team(2023)}]{unsloth}
Michael~Han Daniel~Han and Unsloth team. 2023.
\newblock \href {http://github.com/unslothai/unsloth} {Unsloth}.

\bibitem[{Devlin et~al.(2019)Devlin, Chang, Lee, and Toutanova}]{devlin-etal-2019-bert}
Jacob Devlin, Ming-Wei Chang, Kenton Lee, and Kristina Toutanova. 2019.
\newblock \href {https://doi.org/10.18653/v1/N19-1423} {{BERT}: Pre-training of deep bidirectional transformers for language understanding}.
\newblock In \emph{Proceedings of the 2019 Conference of the North {A}merican Chapter of the Association for Computational Linguistics: Human Language Technologies, Volume 1 (Long and Short Papers)}, pages 4171--4186, Minneapolis, Minnesota. Association for Computational Linguistics.

\bibitem[{Dmonte et~al.(2024)Dmonte, Oruche, Zampieri, Calyam, and Augenstein}]{dmonte2024claim}
Alphaeus Dmonte, Roland Oruche, Marcos Zampieri, Prasad Calyam, and Isabelle Augenstein. 2024.
\newblock Claim verification in the age of large language models: A survey.
\newblock \emph{arXiv preprint arXiv:2408.14317}.

\bibitem[{Dubey et~al.(2024)Dubey, Jauhri, Pandey, Kadian, Al-Dahle, Letman, Mathur, Schelten, Yang, Fan et~al.}]{dubey2024llama}
Abhimanyu Dubey, Abhinav Jauhri, Abhinav Pandey, Abhishek Kadian, Ahmad Al-Dahle, Aiesha Letman, Akhil Mathur, Alan Schelten, Amy Yang, Angela Fan, and 1 others. 2024.
\newblock The llama 3 herd of models.
\newblock \emph{arXiv preprint arXiv:2407.21783}.

\bibitem[{Fung et~al.(2022)Fung, Huang, Nakov, and Ji}]{fung2022battlefront}
Yi~R Fung, Kung-Hsiang Huang, Preslav Nakov, and Heng Ji. 2022.
\newblock The battlefront of combating misinformation and coping with media bias.
\newblock In \emph{Proceedings of the 28th ACM SIGKDD Conference on Knowledge Discovery and Data Mining}, pages 4790--4791.

\bibitem[{Guo et~al.(2022)Guo, Schlichtkrull, and Vlachos}]{guo-etal-2022-survey}
Zhijiang Guo, Michael Schlichtkrull, and Andreas Vlachos. 2022.
\newblock \href {https://doi.org/10.1162/tacl_a_00454} {A survey on automated fact-checking}.
\newblock \emph{Transactions of the Association for Computational Linguistics}, 10:178--206.

\bibitem[{Gupta and Srikumar(2021)}]{gupta-srikumar-2021-x}
Ashim Gupta and Vivek Srikumar. 2021.
\newblock \href {https://doi.org/10.18653/v1/2021.acl-short.86} {{X}-fact: A new benchmark dataset for multilingual fact checking}.
\newblock In \emph{Proceedings of the 59th Annual Meeting of the Association for Computational Linguistics and the 11th International Joint Conference on Natural Language Processing (Volume 2: Short Papers)}, pages 675--682, Online. Association for Computational Linguistics.

\bibitem[{He et~al.(2021)He, Liu, Gao, and Chen}]{he2021deberta}
Pengcheng He, Xiaodong Liu, Jianfeng Gao, and Weizhu Chen. 2021.
\newblock \href {https://openreview.net/forum?id=XPZIaotutsD} {Deberta: Decoding-enhanced bert with disentangled attention}.
\newblock In \emph{International Conference on Learning Representations}.

\bibitem[{Hu et~al.(2022{\natexlab{a}})Hu, Shen, Wallis, Allen-Zhu, Li, Wang, Wang, and Chen}]{hu2022lowrank}
Edward~J. Hu, Yelong Shen, Phillip Wallis, Zeyuan Allen-Zhu, Yuanzhi Li, Shean Wang, Lu~Wang, and Weizhu Chen. 2022{\natexlab{a}}.
\newblock \href {http://dblp.uni-trier.de/db/conf/iclr/iclr2022.html#HuSWALWWC22} {Lora: Low-rank adaptation of large language models.}
\newblock In \emph{ICLR}. OpenReview.net.

\bibitem[{Hu et~al.(2023)Hu, Chen, Li, Guo, Wen, Yu, and Guo}]{hu2023largelanguagemodelsknow}
Xuming Hu, Junzhe Chen, Xiaochuan Li, Yufei Guo, Lijie Wen, Philip~S. Yu, and Zhijiang Guo. 2023.
\newblock \href {https://arxiv.org/abs/2310.05177} {Do large language models know about facts?}
\newblock \emph{Preprint}, arXiv:2310.05177.

\bibitem[{Hu et~al.(2022{\natexlab{b}})Hu, Guo, Wu, Liu, Wen, and Yu}]{hu-etal-2022-chef}
Xuming Hu, Zhijiang Guo, GuanYu Wu, Aiwei Liu, Lijie Wen, and Philip Yu. 2022{\natexlab{b}}.
\newblock \href {https://doi.org/10.18653/v1/2022.naacl-main.246} {{CHEF}: A pilot {C}hinese dataset for evidence-based fact-checking}.
\newblock In \emph{Proceedings of the 2022 Conference of the North American Chapter of the Association for Computational Linguistics: Human Language Technologies}, pages 3362--3376, Seattle, United States. Association for Computational Linguistics.

\bibitem[{Jiang et~al.(2020)Jiang, Yu, Zhou, Chen, Feng, and Yan}]{NEURIPS2020_96da2f59}
Zi-Hang Jiang, Weihao Yu, Daquan Zhou, Yunpeng Chen, Jiashi Feng, and Shuicheng Yan. 2020.
\newblock \href {https://proceedings.neurips.cc/paper/2020/file/96da2f590cd7246bbde0051047b0d6f7-Paper.pdf} {Convbert: Improving bert with span-based dynamic convolution}.
\newblock In \emph{Advances in Neural Information Processing Systems}, volume~33, pages 12837--12848. Curran Associates, Inc.

\bibitem[{Li et~al.(2020)Li, Jiang, Shu, and Liu}]{li2020mmcovidmultilingualmultimodaldata}
Yichuan Li, Bohan Jiang, Kai Shu, and Huan Liu. 2020.
\newblock \href {https://arxiv.org/abs/2011.04088} {Mm-covid: A multilingual and multimodal data repository for combating covid-19 disinformation}.
\newblock \emph{Preprint}, arXiv:2011.04088.

\bibitem[{Liu et~al.(2019)Liu, Ott, Goyal, Du, Joshi, Chen, Levy, Lewis, Zettlemoyer, and Stoyanov}]{liu2019roberta}
Yinhan Liu, Myle Ott, Naman Goyal, Jingfei Du, Mandar Joshi, Danqi Chen, Omer Levy, Mike Lewis, Luke Zettlemoyer, and Veselin Stoyanov. 2019.
\newblock Roberta: A robustly optimized bert pretraining approach.
\newblock \emph{arXiv preprint arXiv:1907.11692}.

\bibitem[{{Mistral AI Team}(2024)}]{mistral}
{Mistral AI Team}. 2024.
\newblock Mistral nemo.
\newblock \url{https://mistral.ai/en/news/mistral-nemo}.
\newblock Accessed: 14-Feb-2025.

\bibitem[{Mohtaj et~al.(2024)Mohtaj, Nizamoglu, Sahitaj, Schmitt, Jakob, and M\"{o}ller}]{Mohtaj}
Salar Mohtaj, Ata Nizamoglu, Premtim Sahitaj, Vera Schmitt, Charlott Jakob, and Sebastian M\"{o}ller. 2024.
\newblock \href {https://doi.org/10.1145/3643491.3660290} {Newspolyml: Multi-lingual european news fake assessment dataset}.
\newblock In \emph{Proceedings of the 3rd ACM International Workshop on Multimedia AI against Disinformation}, MAD '24, page 82–90, New York, NY, USA. Association for Computing Machinery.

\bibitem[{N{\o}rregaard and Derczynski(2021)}]{norregaard-derczynski-2021-danfever}
Jeppe N{\o}rregaard and Leon Derczynski. 2021.
\newblock \href {https://aclanthology.org/2021.nodalida-main.47/} {{D}an{FEVER}: claim verification dataset for {D}anish}.
\newblock In \emph{Proceedings of the 23rd Nordic Conference on Computational Linguistics (NoDaLiDa)}, pages 422--428, Reykjavik, Iceland (Online). Link{\"o}ping University Electronic Press, Sweden.

\bibitem[{Pelrine et~al.(2023)Pelrine, Imouza, Thibault, Reksoprodjo, Gupta, Christoph, Godbout, and Rabbany}]{pelrine-etal-2023-towards}
Kellin Pelrine, Anne Imouza, Camille Thibault, Meilina Reksoprodjo, Caleb Gupta, Joel Christoph, Jean-Fran{\c{c}}ois Godbout, and Reihaneh Rabbany. 2023.
\newblock \href {https://doi.org/10.18653/v1/2023.emnlp-main.395} {Towards reliable misinformation mitigation: Generalization, uncertainty, and {GPT}-4}.
\newblock In \emph{Proceedings of the 2023 Conference on Empirical Methods in Natural Language Processing}, pages 6399--6429, Singapore. Association for Computational Linguistics.

\bibitem[{Pikuliak et~al.(2023)Pikuliak, Srba, Moro, Hromadka, Smole{\v{n}}, Meli{\v{s}}ek, Vykopal, Simko, Podrou{\v{z}}ek, and Bielikova}]{pikuliak-etal-2023-multilingual}
Mat{\'u}{\v{s}} Pikuliak, Ivan Srba, Robert Moro, Timo Hromadka, Timotej Smole{\v{n}}, Martin Meli{\v{s}}ek, Ivan Vykopal, Jakub Simko, Juraj Podrou{\v{z}}ek, and Maria Bielikova. 2023.
\newblock \href {https://doi.org/10.18653/v1/2023.emnlp-main.1027} {Multilingual previously fact-checked claim retrieval}.
\newblock In \emph{Proceedings of the 2023 Conference on Empirical Methods in Natural Language Processing}, pages 16477--16500, Singapore. Association for Computational Linguistics.

\bibitem[{Poliak et~al.(2018)Poliak, Naradowsky, Haldar, Rudinger, and Van~Durme}]{poliak-etal-2018-hypothesis}
Adam Poliak, Jason Naradowsky, Aparajita Haldar, Rachel Rudinger, and Benjamin Van~Durme. 2018.
\newblock \href {https://doi.org/10.18653/v1/S18-2023} {Hypothesis only baselines in natural language inference}.
\newblock In \emph{Proceedings of the Seventh Joint Conference on Lexical and Computational Semantics}, pages 180--191, New Orleans, Louisiana. Association for Computational Linguistics.

\bibitem[{Scheufele and Krause(2019)}]{scheufele2019science}
Dietram~A Scheufele and Nicole~M Krause. 2019.
\newblock Science audiences, misinformation, and fake news.
\newblock \emph{Proceedings of the National Academy of Sciences}, 116(16):7662--7669.

\bibitem[{Shahi and Nandini(2020)}]{shahi_nandini_2020}
Gautam~Kishore Shahi and Durgesh Nandini. 2020.
\newblock \href {https://doi.org/10.36190/2020.14} {\emph{FakeCovid- A Multilingual Cross-domain Fact Check News Dataset for COVID-19}}.
\newblock ICWSM.

\bibitem[{Shahi et~al.(2021)Shahi, Stru{\ss}, and Mandl}]{shahi2021overview}
Gautam~Kishore Shahi, Julia~Maria Stru{\ss}, and Thomas Mandl. 2021.
\newblock Overview of the clef-2021 checkthat! lab task 3 on fake news detection.
\newblock \emph{Working Notes of CLEF}.

\bibitem[{Singhal et~al.(2024)Singhal, Law, Kassner, Gupta, Duan, Damle, and Li}]{singhal-etal-2024-multilingual}
Aryan Singhal, Thomas Law, Coby Kassner, Ayushman Gupta, Evan Duan, Aviral Damle, and Ryan~Luo Li. 2024.
\newblock \href {https://aclanthology.org/2024.nlp4pi-1.2} {Multilingual fact-checking using {LLM}s}.
\newblock In \emph{Proceedings of the Third Workshop on NLP for Positive Impact}, pages 13--31, Miami, Florida, USA. Association for Computational Linguistics.

\bibitem[{Touvron et~al.(2023)Touvron, Martin, Stone, Albert, Almahairi, Babaei, Bashlykov, Batra, Bhargava, Bhosale, Bikel, Blecher, Ferrer, Chen, Cucurull, Esiobu, Fernandes, Fu, Fu, Fuller, Gao, Goswami, Goyal, Hartshorn, Hosseini, Hou, Inan, Kardas, Kerkez, Khabsa, Kloumann, Korenev, Koura, Lachaux, Lavril, Lee, Liskovich, Lu, Mao, Martinet, Mihaylov, Mishra, Molybog, Nie, Poulton, Reizenstein, Rungta, Saladi, Schelten, Silva, Smith, Subramanian, Tan, Tang, Taylor, Williams, Kuan, Xu, Yan, Zarov, Zhang, Fan, Kambadur, Narang, Rodriguez, Stojnic, Edunov, and Scialom}]{touvron2023llama2openfoundation}
Hugo Touvron, Louis Martin, Kevin Stone, Peter Albert, Amjad Almahairi, Yasmine Babaei, Nikolay Bashlykov, Soumya Batra, Prajjwal Bhargava, Shruti Bhosale, Dan Bikel, Lukas Blecher, Cristian~Canton Ferrer, Moya Chen, Guillem Cucurull, David Esiobu, Jude Fernandes, Jeremy Fu, Wenyin Fu, and 49 others. 2023.
\newblock \href {https://arxiv.org/abs/2307.09288} {Llama 2: Open foundation and fine-tuned chat models}.
\newblock \emph{Preprint}, arXiv:2307.09288.

\bibitem[{Ullrich et~al.(2023)Ullrich, Drchal, Rýpar, Vincourová, and Moravec}]{Ullrich_2023}
Herbert Ullrich, Jan Drchal, Martin Rýpar, Hana Vincourová, and Václav Moravec. 2023.
\newblock \href {https://doi.org/10.1007/s10579-023-09654-3} {Csfever and ctkfacts: acquiring czech data for fact verification}.
\newblock \emph{Language Resources and Evaluation}, 57(4):1571–1605.

\bibitem[{Wang(2017)}]{wang-2017-liar}
William~Yang Wang. 2017.
\newblock \href {https://doi.org/10.18653/v1/P17-2067} {{\textquotedblleft}liar, liar pants on fire{\textquotedblright}: A new benchmark dataset for fake news detection}.
\newblock In \emph{Proceedings of the 55th Annual Meeting of the Association for Computational Linguistics (Volume 2: Short Papers)}, pages 422--426, Vancouver, Canada. Association for Computational Linguistics.

\bibitem[{Wang et~al.(2024)Wang, Zhang, and Rajtmajer}]{wang2024monolingualmultilingualmisinformationdetection}
Xinyu Wang, Wenbo Zhang, and Sarah Rajtmajer. 2024.
\newblock \href {https://arxiv.org/abs/2410.18390} {Monolingual and multilingual misinformation detection for low-resource languages: A comprehensive survey}.
\newblock \emph{Preprint}, arXiv:2410.18390.

\bibitem[{Wei et~al.(2022)Wei, Wang, Schuurmans, Bosma, Ichter, Xia, Chi, Le, and Zhou}]{wei}
Jason Wei, Xuezhi Wang, Dale Schuurmans, Maarten Bosma, Brian Ichter, Fei Xia, Ed~H. Chi, Quoc~V. Le, and Denny Zhou. 2022.
\newblock Chain-of-thought prompting elicits reasoning in large language models.
\newblock In \emph{Proceedings of the 36th International Conference on Neural Information Processing Systems}, NIPS '22, Red Hook, NY, USA. Curran Associates Inc.

\bibitem[{Xue et~al.(2021)Xue, Constant, Roberts, Kale, Al-Rfou, Siddhant, Barua, and Raffel}]{xue-etal-2021-mt5}
Linting Xue, Noah Constant, Adam Roberts, Mihir Kale, Rami Al-Rfou, Aditya Siddhant, Aditya Barua, and Colin Raffel. 2021.
\newblock \href {https://doi.org/10.18653/v1/2021.naacl-main.41} {m{T}5: A massively multilingual pre-trained text-to-text transformer}.
\newblock In \emph{Proceedings of the 2021 Conference of the North American Chapter of the Association for Computational Linguistics: Human Language Technologies}, pages 483--498, Online. Association for Computational Linguistics.

\bibitem[{Yang et~al.(2024)Yang, Yang, Zhang, Hui, Zheng, Yu, Li, Liu, Huang, Wei et~al.}]{yang2024qwen2}
An~Yang, Baosong Yang, Beichen Zhang, Binyuan Hui, Bo~Zheng, Bowen Yu, Chengyuan Li, Dayiheng Liu, Fei Huang, Haoran Wei, and 1 others. 2024.
\newblock Qwen2. 5 technical report.
\newblock \emph{arXiv preprint arXiv:2412.15115}.

\bibitem[{Zhang et~al.(2024)Zhang, Guo, and Vlachos}]{zhang-etal-2024-need}
Caiqi Zhang, Zhijiang Guo, and Andreas Vlachos. 2024.
\newblock \href {https://aclanthology.org/2024.emnlp-main.113} {Do we need language-specific fact-checking models? the case of {C}hinese}.
\newblock In \emph{Proceedings of the 2024 Conference on Empirical Methods in Natural Language Processing}, pages 1899--1914, Miami, Florida, USA. Association for Computational Linguistics.

\end{thebibliography}

\appendix

\section{Model Implementation Details}
\label{app:models}

For our experimental evaluation, we used the following model versions:

\textbf{XLM-R base}. We used \textit{FacebookAI/xlm-roberta-base} (270 million parameters) model from Hugging Face, which has been pre-trained on text in 100 languages. The model was tested in two configurations: (1) with frozen parameters and only the classification head fine-tuned, and (2) with full fine-tuning of all parameters.

\textbf{mT5 base}. We employed the \textit{google/mt5-base} model (580 million parameters) from Hugging Face, which follows an encoder-decoder architecture and has been pre-trained on multilingual text.

\textbf{Llama 3.1 8B}. We used the instruction-tuned version of Llama 3.1 with 8 billion parameters. This model officially supports seven languages: French, German, Hindi, Italian, Portuguese, Spanish, and Thai, in addition to English.

\textbf{Qwen 2.5 7B}. We employed the instruction-tuned Qwen 2.5 model with 7 billion parameters. This model supports 29 languages and has demonstrated strong performance in both English and multilingual tasks.

\textbf{Mistral Nemo 12B}. We used the Mistral Nemo model with 12 billion parameters. This model supports 11 languages: English, French, German, Spanish, Italian, Portuguese, Chinese, Japanese, Korean, Arabic, and Hindi.

All experiments with LLMs were conducted using the Unsloth library \citep{unsloth} to efficiently implement and optimize the fine-tuning and inference processes, ensuring faster training times and reduced memory usage without compromising model performance. 

\section{Details on X-Fact}
\label{app:claim}

\begin{figure}[h]
  \includegraphics[width=\columnwidth]{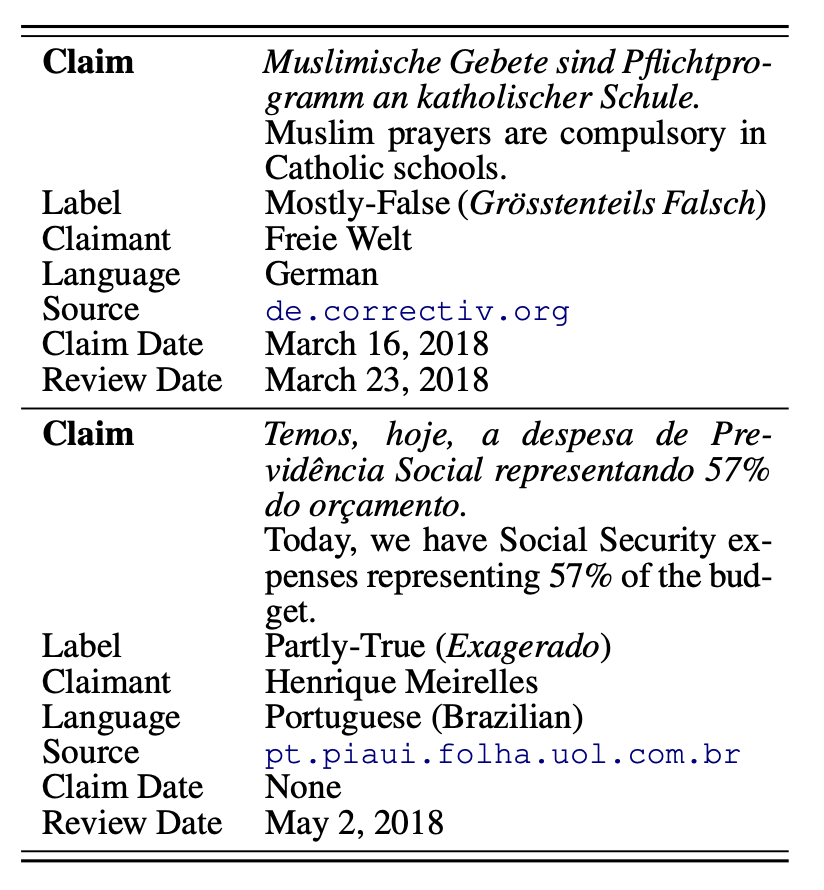}
  \caption{Details of the X-Fact dataset. Examples from X-Fact as presented in the original paper by \citet{gupta-srikumar-2021-x}. For reference, translations are also shown.}
  \label{fig:claim}
\end{figure}

For examples from the X-Fact dataset, please refer to the Figure \ref{fig:claim}.

% \section{Language Family Distribution in X-Fact}
% \label{app:families}

% \begin{figure*}[h]
%   \includegraphics[width=\textwidth]{latex/families.png}
%   \caption{Details of the X-FACT dataset. The dataset contains text data in 25 typologically diverse languages across 11 language families. IE: denotes Indo-European \citep{gupta-srikumar-2021-x}.}
%   \label{fig:families}
% \end{figure*}

\section{Hyperparameter Details}
\label{app:hypers}
\subsection{Small Language Models}

For our small language models (XLM-R and mT5), we employed Bayesian hyperparameter optimization through Weights\&Biases, conducting 90 sweeps for the classification head approach and 60 sweeps each for the full fine-tuning experiments. We used an AdamW optimizer with a polynomial learning rate scheduler. To prevent overfitting, we implemented early stopping. Table \ref{tab:slm_hypers} shows the key hyperparameter values for each model variant.

\begin{table}[h]
  \centering
  \begin{tabular}{l|c|c}
    \hline
    \textbf{Model} & \textbf{Learning Rate} & \textbf{Batch Size} \\
    \hline
    XLM-R frozen & 5.7e-04 & 8 \\
    XLM-R & 1.82e-05 & 6 \\
    mT5 & 2.2e-05 & 8 \\
    \hline
  \end{tabular}
  \caption{Key hyperparameter values for SLMs.}
  \label{tab:slm_hypers}
\end{table}

\subsection{Large Language Models}

For large language models, we used parameter-efficient fine-tuning with LoRA. Through systematic experimentation, we identified optimal LoRA configurations with a rank of 16 and adapter alpha of 32. We targeted both attention components (query, key, value, and output projections) and feed-forward layers (gate projections and up/down projections).

Lower rank values (r = 2, 4, 8) and alpha values (8, 16) produced inferior results, while increasing these parameters beyond our chosen values (r > 16, alpha > 32) provided negligible performance gains while substantially increasing memory requirements.

For prompt engineering, we tested various temperature settings and found that temperatures between 0.3 and 0.5 provided the best balance between confident predictions and appropriate uncertainty handling. Lower temperatures led to overly deterministic outputs that failed to capture nuanced veracity judgments, while higher temperatures resulted in inconsistent classifications.

All LLM experiments were conducted using 4-bit quantization to enable efficient processing on GPUs while maintaining performance. 

\section{Performance Comparison across Models}
\label{app:all}

\begin{figure}[h]
  \includegraphics[width=\columnwidth]{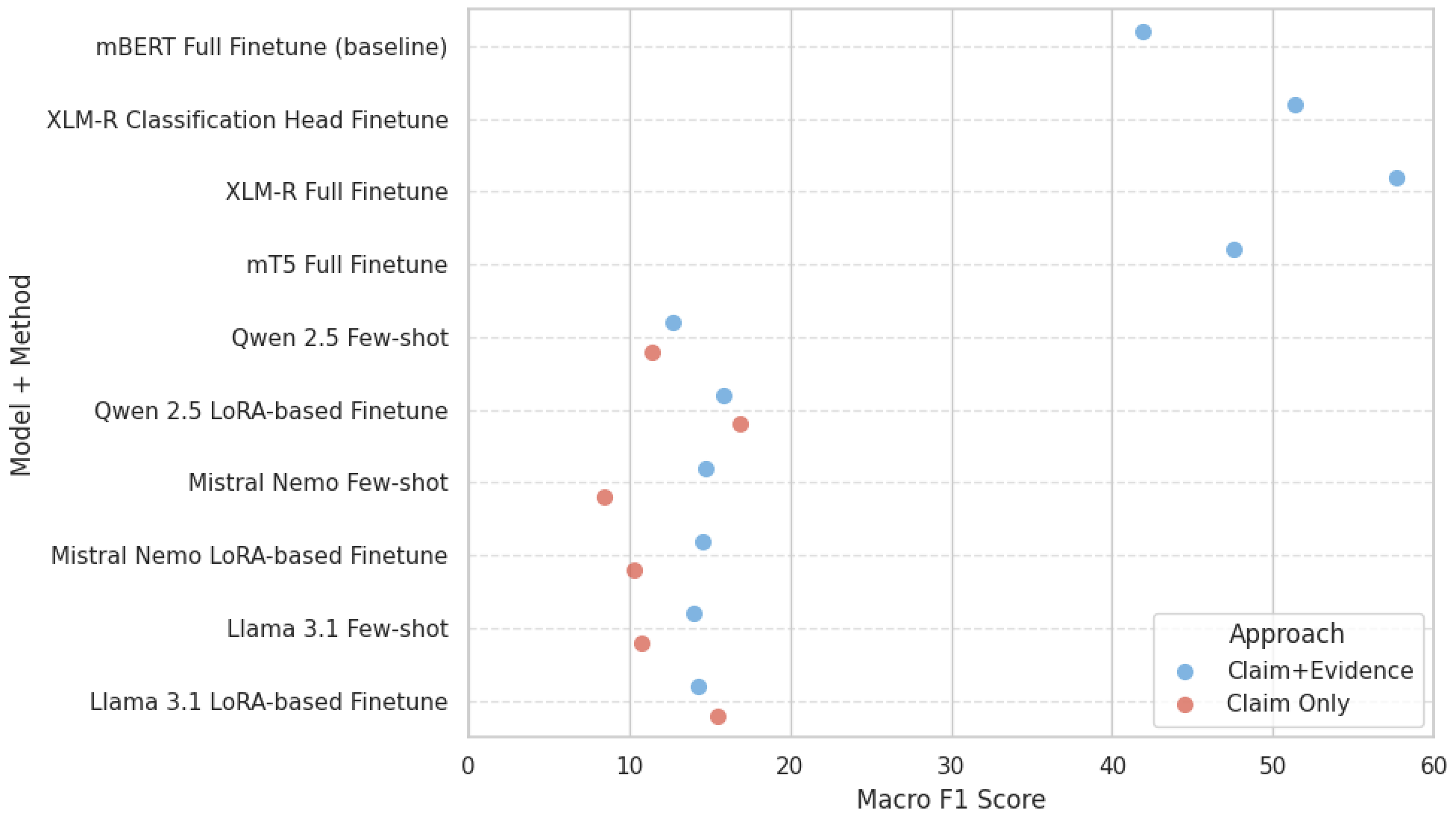}
  \caption{Macro-F1 scores across test subset by model.}
  \label{fig:test}
\end{figure}

\begin{figure}[h]
  \includegraphics[width=\columnwidth]{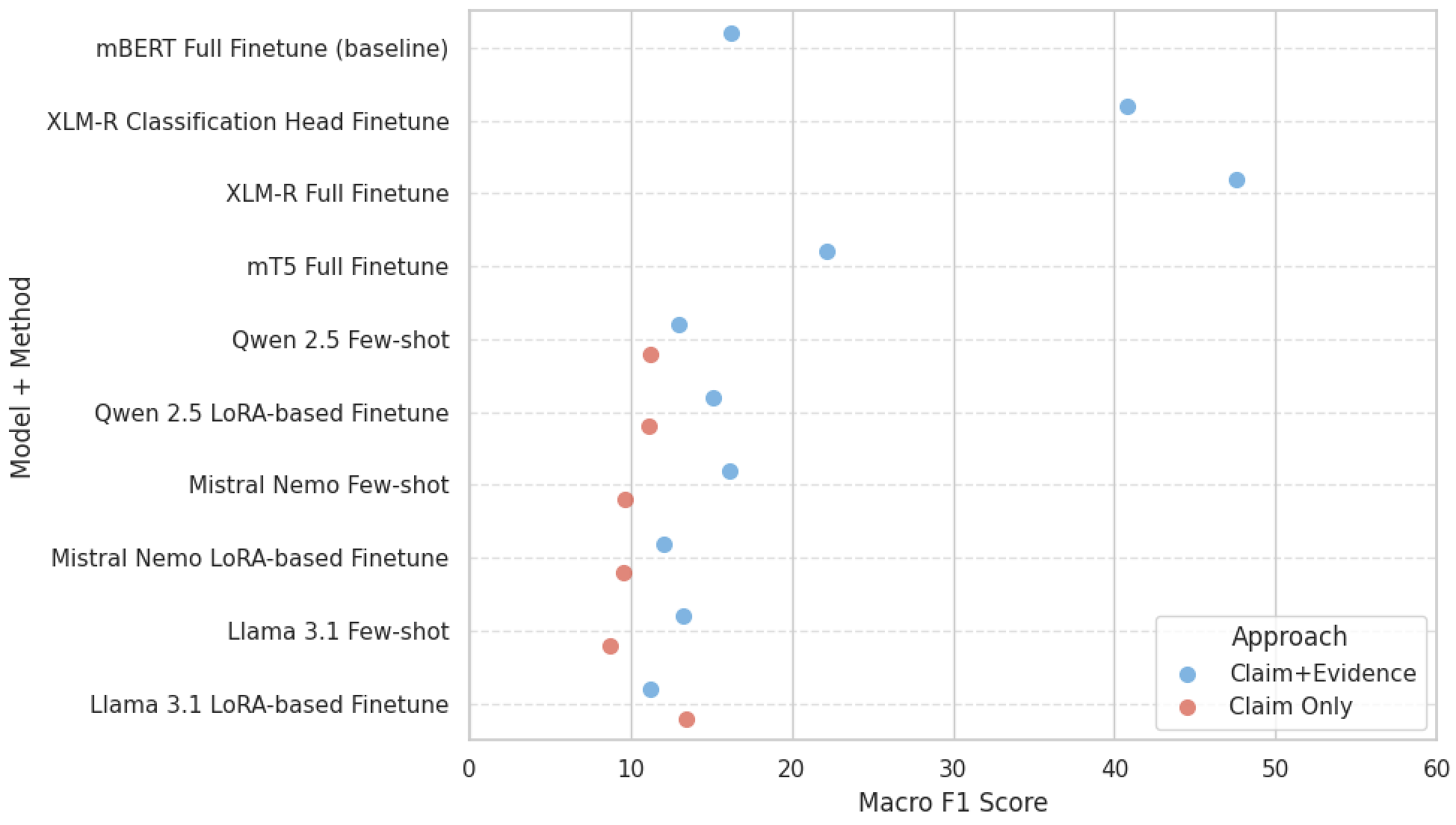}
  \caption{Macro-F1 scores across OOD subset by model.}
  \label{fig:ood}
\end{figure}

\begin{figure}[h]
  \includegraphics[width=\columnwidth]{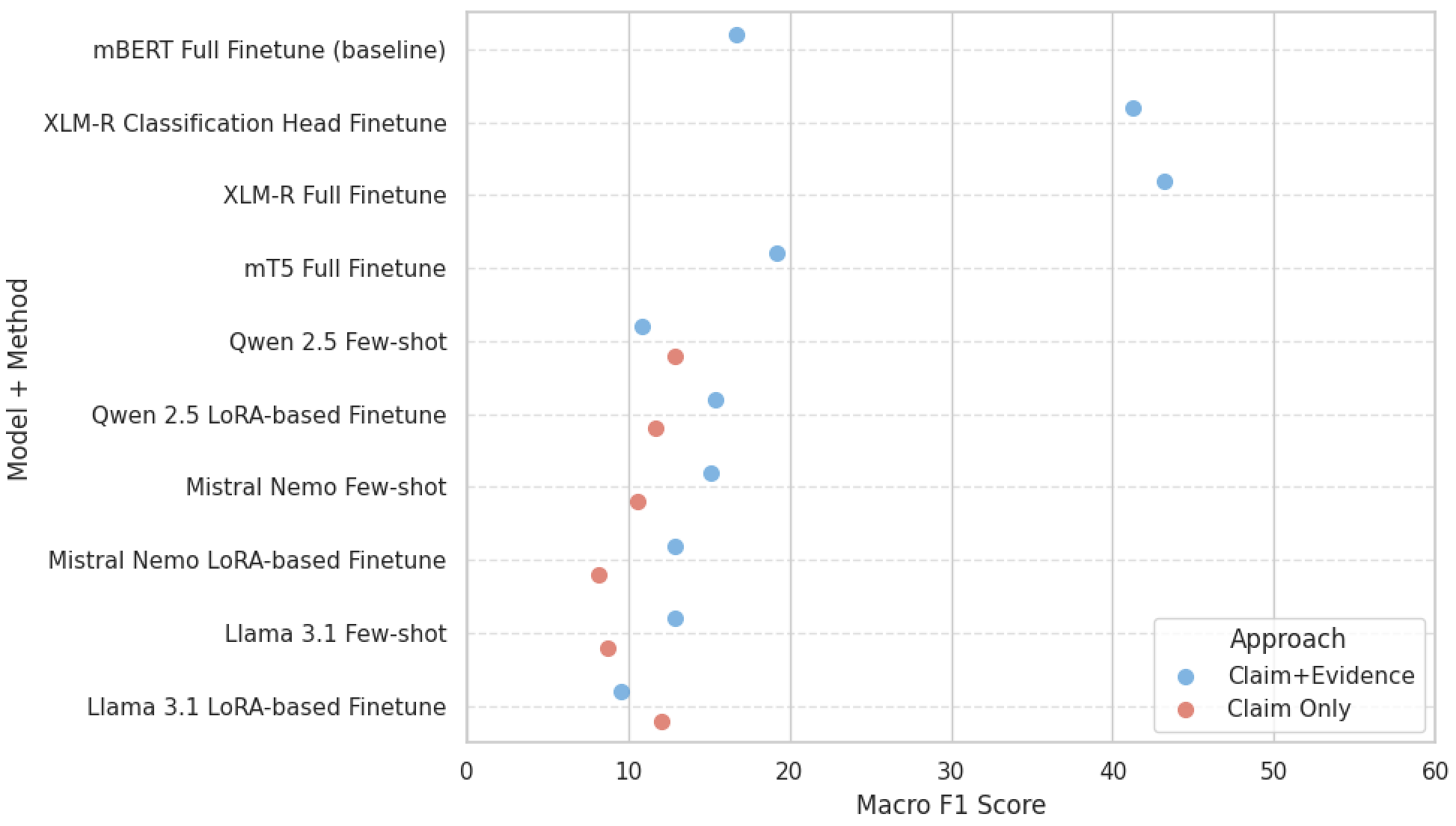}
  \caption{Macro-F1 scores across zero-shot subset by model.}
  \label{fig:zero}
\end{figure}

\section{Prompt Template}
\begin{figure*}[h]
  \includegraphics[width=\textwidth]{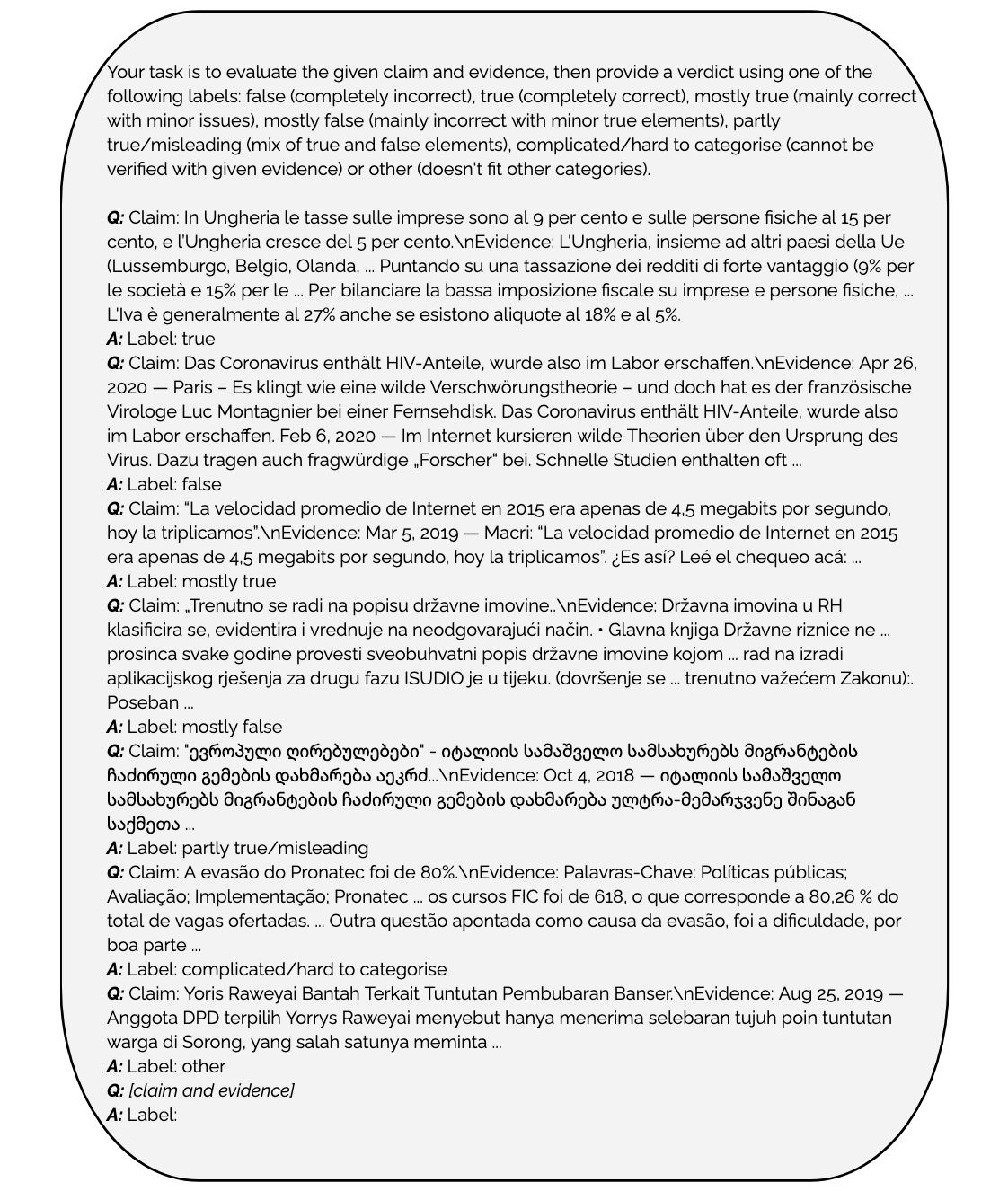}
  \caption{Prompt template used for LLMs.}
  \label{fig:prompt}
\end{figure*}

In Figure \ref{fig:prompt} we provide a prompt template used to instruct LLMs.
\label{app:prompt}

\section{LLMs Performance Comparison Visualizations}
\label{app:set}

\begin{figure*}[h]
  \includegraphics[width=\textwidth]{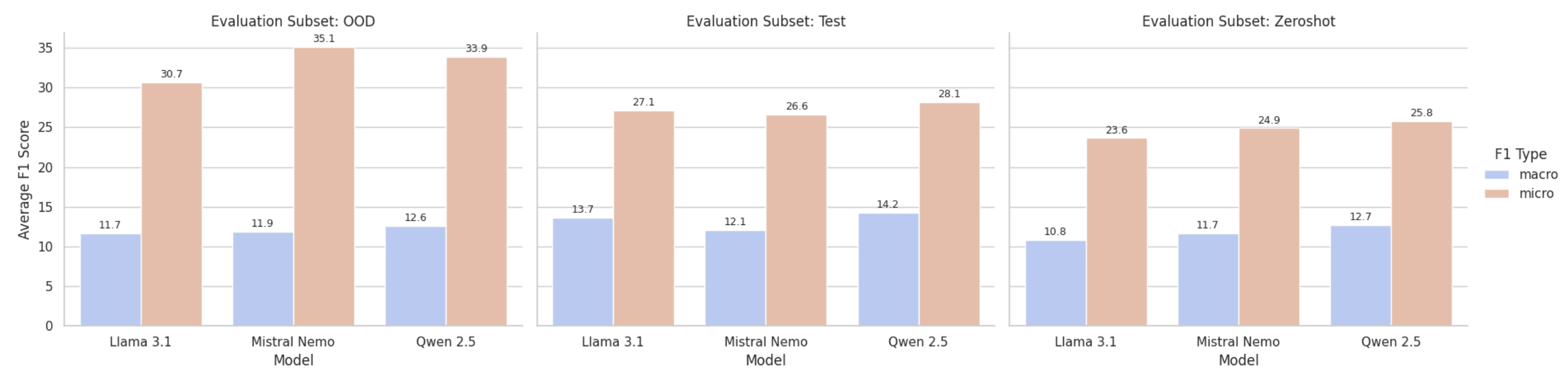}
  \caption{Comparison of macro- and micro F1 scores across LLMs, evaluation subsets, and training methods.}
  \label{fig:set}
\end{figure*}

In Figure \ref{fig:set} we provide a comparison of macro- and micro F1 scores across LLMs, evaluation subsets, and training methods.

\section{LLMs Performance Summary Table}
\label{app:consol}

In Table \ref{tab:combined_llm_performance} we present a combined comparison of macro-F1 scores for all evaluated models using claim+evidence configurations across the three evaluation subsets (Test, OOD, Zero-shot). This table extracts the claim+evidence results from Table 4 and combines them with the small language model performance to facilitate direct performance comparison.

\begin{table*}[htbp]
\centering
\begin{tabular}{l|c|c|c}
\toprule
\textbf{Method} & \textbf{Test} & \textbf{OOD} & \textbf{Zero-shot} \\
\midrule
mBERT (SLM) & 41.9 & 16.2 & 16.7 \\
XLM-R frozen (SLM) & 51.4 & 40.8 & 41.3 \\
XLM-R (SLM) & \textbf{57.7} & \textbf{47.6} & \textbf{43.2} \\
mT5 (SLM) & 47.6 & 22.2 & 19.2 \\
\midrule
Qwen 2.5 Few-shot (LLM) & 12.7 & 13.0 & 10.9 \\
Qwen 2.5 LoRA (LLM) & \textbf{15.9} & \textbf{15.1} & \textbf{15.4} \\
\midrule
Mistral Nemo Few-shot (LLM) & \textbf{14.8} & \textbf{16.1} & \textbf{15.1} \\
Mistral Nemo LoRA (LLM) & 14.6 & 12.1 & 12.9 \\
\midrule
Llama 3.1 Few-shot (LLM) & 14.0 & \textbf{13.3} & \textbf{12.9} \\
Llama 3.1 LoRA (LLM) & \textbf{14.3} & 11.2 & 9.6 \\
\bottomrule
\end{tabular}
\caption{Macro-F1 performance comparison across evaluation subsets for claim+evidence configurations. Bold values indicate the highest macro-F1 score for each LLM model across the two training methods (Few-shot vs LoRA). SLMs results included for reference.}
\label{tab:combined_llm_performance}
\end{table*}

\section{Micro- and Macro-F1 Scores Comparison across LLMs}
\label{app:f1}

\begin{figure*}[h]
  \includegraphics[width=\textwidth]{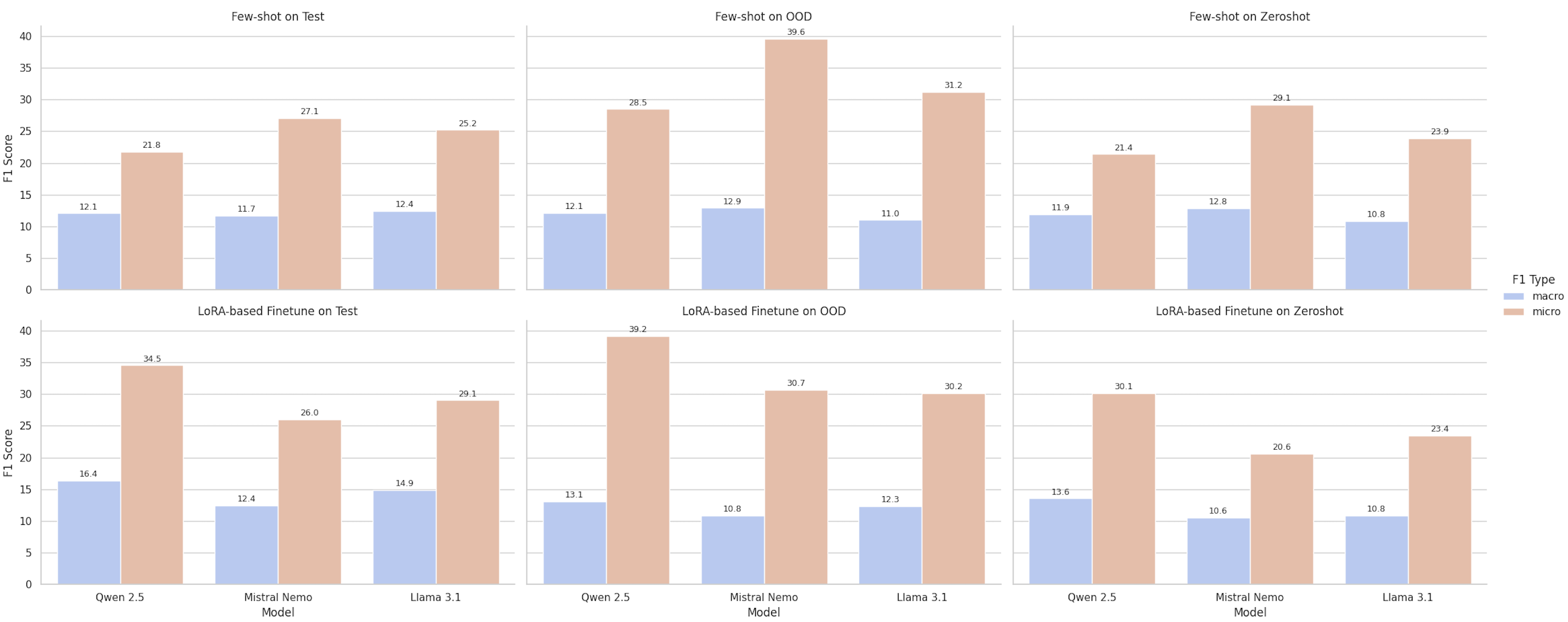}
  \caption{Comparison of average macro- and micro-F1 scores across LLMs for each evaluation subset.}
  \label{fig:f1}
\end{figure*}

In Figure \ref{fig:f1} we provide a comparison of average macro- and micro-F1 scores across LLMs for each evaluation subset.

\end{document}